\documentclass[runningheads]{llncs}
\usepackage{graphicx}

\usepackage{tikz}
\usepackage{comment}
\usepackage{amsmath,amssymb} 
\usepackage{color}

\usepackage{booktabs}
\usepackage{caption}
\usepackage{subcaption}

\usepackage{bbm}
\usepackage{multirow}
\usepackage{xcolor}
\usepackage{pifont}
\newcommand{\xmark}{\ding{55}}
\newcommand{\cmark}{\ding{51}}
\usepackage{algorithm}
\usepackage{algorithmic}

\usepackage[accsupp]{axessibility}  

\usepackage[width=122mm,left=12mm,paperwidth=146mm,height=193mm,top=12mm,paperheight=217mm]{geometry}

\usepackage[capitalize]{cleveref}
\crefname{section}{Sec.}{Secs.}
\Crefname{section}{Section}{Sections}
\Crefname{table}{Table}{Tables}
\crefname{table}{Tab.}{Tabs.}

\begin{document}
\pagestyle{headings}
\mainmatter
\def\ECCVSubNumber{4309}  

\title{From One to Many: Dynamic Cross Attention Networks for LiDAR and Camera Fusion} 


%
\author{Rui Wan\inst{1}\and
Shuangjie Xu\inst{1,2} \and Wei Wu\inst{1} \and Xiaoyi Zou\inst{1} \and Tongyi Cao\inst{1}}
%
%
\institute{DEEPROUTE.AI \and Hong Kong University of Science and Technology
\email{\{ruiwan,weiwu,xiaoyizou,tongyicao\}@deeproute.ai sxubj@connect.ust.hk}}

\maketitle

\begin{abstract}
LiDAR and cameras are two complementary sensors for 3D perception in autonomous driving. LiDAR point clouds have accurate spatial and geometry information, while RGB images provide textural and color data for context reasoning. To exploit LiDAR and cameras jointly, existing fusion methods tend to align each 3D point to only one projected image pixel based on calibration, namely \textit{one-to-one} mapping. However, the performance of these approaches highly relies on the calibration quality, which is sensitive to the temporal and spatial synchronization of sensors. Therefore, we propose a Dynamic Cross Attention (DCA) module with a novel \textit{one-to-many} cross-modality mapping that learns multiple offsets from the initial projection towards the neighborhood and thus develops tolerance to calibration error. Moreover, a \textit{dynamic query enhancement} is proposed to perceive the model-independent calibration, which further strengthens DCA's tolerance to the initial misalignment. The whole fusion architecture named Dynamic Cross Attention Network (DCAN) exploits multi-level image features and adapts to multiple representations of point clouds, which allows DCA to serve as a plug-in fusion module. Extensive experiments on nuScenes and KITTI prove DCA's effectiveness. The proposed DCAN outperforms state-of-the-art methods on the nuScenes detection challenge.
\keywords{LiDAR and camera fusion, 3D detection}
\end{abstract}

\section{Introduction}
 \label{sec:intro}

3D object detection has been a crucial task for 3D perception in autonomous driving. Although point clouds based approaches have achieved great progress in this task, its sparsity is a challenge for small-object and long-range perception. The camera contains dense textural and color information but lacks depth for accurate 3D localization. Therefore, an effective fusion of LiDAR and cameras is required for accurate and robust 3D perception.

\begin{figure}
    \vspace {5mm}
    \centering
    \includegraphics[width=1\textwidth]{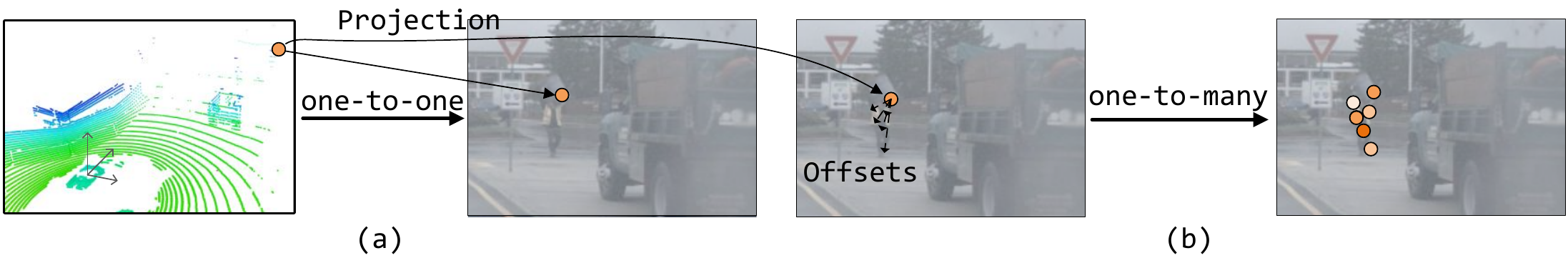}
    \caption{\textit{One-to-one} (a) v.s. \textit{one-to-many} (b)}
    \label{fig:one-to-many}
\end{figure}

Currently, the main challenge of fusion lies in the misalignment of LiDAR and cameras. Existing research aligns these two modalities by proposal-wise and point-wise methods. 
The early proposal-wise methods~\cite{MV3D,AVOD,zhu2021cross} fuse 2D and 3D features from corresponding regions of interest(ROI), which is not only time-consuming because of the large number of proposals but also imprecise due to the gapsetween perspectives and orientation.
Recent point-wise methods~\cite{sindagi2019mvx,vora2020pointpainting,huang2020epnet,liang2019multi,yoo20203d} align cross-modality feature pairs by projecting 3D points to the image plane. Such alignment is \textit{one-to-one} because a 3D point, grid, or voxel corresponds to only one image pixel based on the calibration. Although some point-wise methods such as PointPainting~\cite{vora2020pointpainting} have shown clear advantages against the LiDAR-only methods, they are limited by the nature of calibration. Classical calibration methods optimize the extrinsic matrices in an iterative and dynamic manner~\cite{lcct,scaramuzza2007extrinsic,geiger2012automatic}. As a result of such dynamic optimization, the calibration matrices and the consequent alignment of feature pairs are also dynamic. However, the \textit{one-to-one} mapping is fixed by the calibration that is independent of the model, which makes the point-wise methods unaware of and thus weak against the inevitable calibration error caused by the temporal and spatial synchronization of sensors.

Attempts have been made to learn the relationships of cross-modality features by the network itself and get rid of the calibration error from fusion. Transfuser~\cite{prakash2021multi} is the pioneer that exploits the self-attention module of Transformer~\cite{vaswani2017attention,carion2020end} to capture global feature relationships of LiDAR and cameras, which makes the model independent of calibration. However, there are two main obstacles to its application in large-scale outdoor scenarios. First, the model scalability is rather limited due to the high computational complexity of the standard global attention. Second, the features from multiple modalities are simply stacked as the input sequences without any prior conditions, which makes the model hard to converge to the optimal alignment.

To model the dynamic image contexts, learned offsets that stretch the convolution filters have proved its generalized effectiveness in DCN~\cite{dai2017deformable}, DCNv2~\cite{zhu2019deformablev21}, and Deformable DETR~\cite{zhu2020deformable}. Inspired by its success, we propose a Dynamic Cross Attention (DCA) module that is the first to use a \textit{one-to-many} mapping to model the dynamic alignment of LiDAR and cameras. In contrast to the \textit{one-to-one} mapping, DCA takes the calibration as the initial clue and learns multiple offsets and weights towards the neighboring space as well as the adjacent feature levels, as in Fig.~\ref{fig:one-to-many}. On the one hand, such \textit{one-to-many} feature mapping adaptively learns the local contexts and thus can tolerate a level of calibration error. On the other hand, the \textit{one-to-many} mapping degenerates to \textit{one-to-one} when the number of offsets is zero, and thus covers it. Besides, we propose a \textit{dynamic query enhancement} (DQE) that incorporates the initial \textit{one-to-one} feature pairs into the offset prediction. Such feature pairs contain both 3D geometry and initial image neighbor contexts, serving as an informative guide to the optimal alignment and thus enhancing DCA's tolerance to the initial misalignment. Ablation study shows DCA's robustness against the calibration error.

\begin{figure}[th!]
    \centering
    \includegraphics[width=0.65\textwidth]{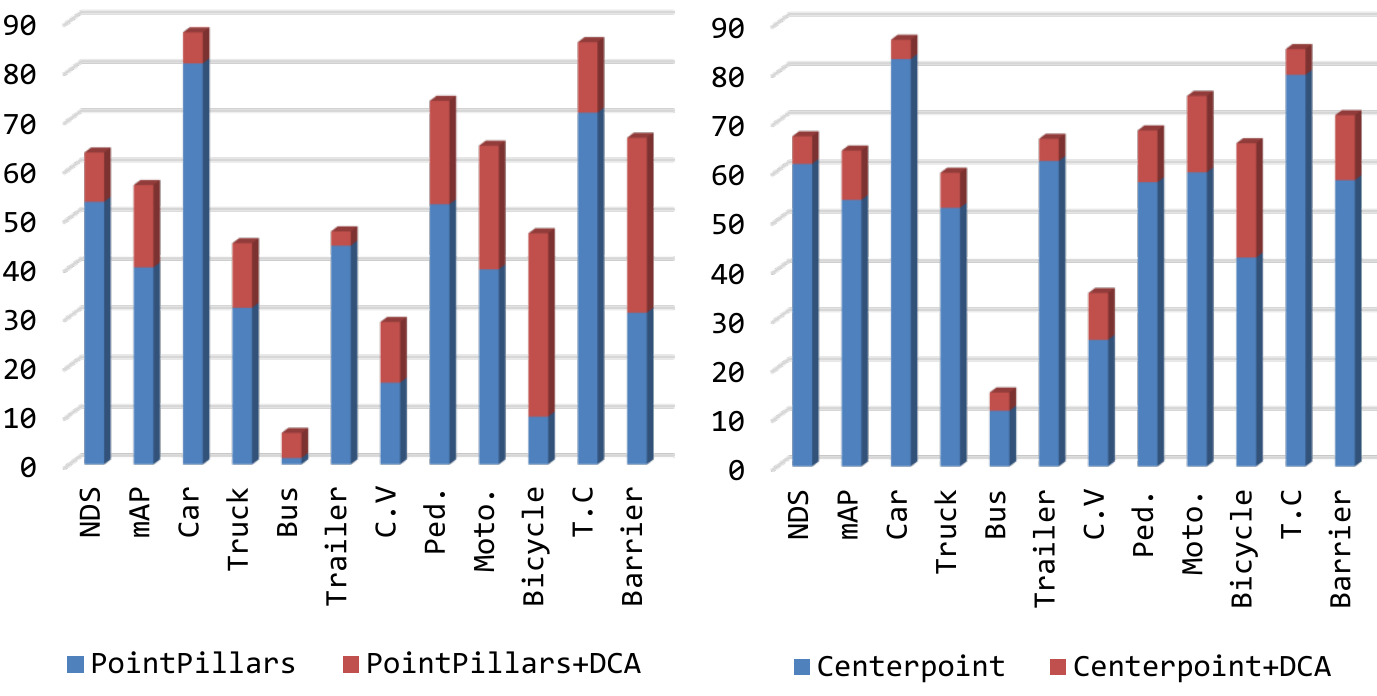}
    \caption{DCA as a plug-in on nuScenes. We draw NDS, mAP, and mAP for each class in \%.}
    \label{fig:improve}
\end{figure}

The whole network, named Dynamic Cross Attention Network (DCAN), consists of the common image and LiDAR branches and the proposed Dynamic Cross Attention (DCA) module. Compared with some fusion methods that only take a single level of image feature for fusion and require costly 2D segmentation annotations, DCAN takes advantage of multi-level image features for fusion and only needs more accessible 2D detection annotations. The choice of the LiDAR branch is flexible as point-based, grid-based, and voxel-based methods can all be easily adapted to our DCA module. We also adopt a joint-training strategy for better feature learning. Extensive experiments on nuScenes~\cite{nuscenes2019} and KITTI~\cite{KITTI} prove DCA's ability to serve as a plug-in fusion module, as in Fig.~\ref{fig:improve}.  Moreover, 
DCAN outperforms published methods for the nuScenes detection challenge. Our main contributions are as followed:

\begin{itemize}
    \item We propose a novel \textit{one-to-many} feature mapping to model the dynamic cross-modality alignment, which adaptively learns the local contexts and thus develops tolerance to the calibration error.
    \item We propose a \textit{dynamic query enhancement} to perceive the model-independent calibration and provide a stronger guide against the misalignment.
    \item We propose a flexible architecture that exploits multi-level image features and adapts to varied representations of LiDAR with the plug-in fusion module.
    \item Extensive experiments on nuScenes and KITTI and the SOTA performance on nuScenes prove the effectiveness of our fusion module and architecture. 
\end{itemize}


\section{Related Work}

\textbf{3D object detection from point clouds.}
Current 3D object detection methods adopt points, grids, and voxels as point cloud representations. Point-based methods including PointRCNN~\cite{shi2019pointrcnn} and 3DSSD~\cite{yang20203dssd} take raw point clouds as inputs and generate boxes based on each point. 
Grid-based and voxel-based methods organize point clouds into regular forms that not only are faster to process especially with the proposal of sparse convolutions~\cite{liu2015sparse,graham20183d} but also can adopt advanced 2D backbones and structures. PIXOR~\cite{yang2018pixor}, PointPillar~\cite{lang2019pointpillars}, and HVNet~\cite{ye2020hvnet} encode grid features and perform detection on BEV maps. VoxelNet~\cite{zhou2018voxelnet}, SECOND~\cite{yan2018second}, and Centerpoint~\cite{yin2021center} exploit sparse voxels and conduct convolutions only on the non-empty voxels, which increases runtime efficiency while effectively encodes the 3D geometry and spatial information.

\noindent\textbf{LiDAR and camera fusion.}
Current LiDAR and camera fusion methods can be divided to proposal-wise and point-wise methods. Proposal-wise methods including MV3D~\cite{MV3D}, AVOD~\cite{AVOD} and~\cite{zhu2021cross} generate and align proposals in 3D and 2D spaces and adopt fully connected layers or ROI pooling for fusion, which leads to heavy computation load given a large number of proposals. Point-wise methods generate cross-modality feature pairs based on calibration. MVXNet~\cite{sindagi2019mvx} and PointPainting~\cite{vora2020pointpainting} concatenate a point cloud feature with the paired image feature or semantic prediction for fusion. EPNet~\cite{huang2020epnet} proposes an LI-Fusion module that learns a weight map for each feature pair. Methods including~\cite{liang2019multi}, 3D-CVF~\cite{yoo20203d} and PointAugmenting~\cite{wang2021pointaugmenting} construct BEV maps of image features to fuse with point cloud BEV features. Such \textit{one-to-one} mapping is fixed by the model-independent calibration, making the fusion alignment weak against the calibration error.

Since ground-truth object sampling augmentation (\textit{i.e.} GT-Aug) has been proved negligible in 2D and 3D classification and object detection tasks ~\cite{yun2019cutmix,bochkovskiy2020yolov4,yan2018second,shi2019pointrcnn}, some fusion methods~\cite{zhang2020exploring,wang2021pointaugmenting} propose GT-Aug that copies object samples from both point clouds and images and paste them during training to boost the performance. However, such augmentation depends on occlusion relationships and handcraft rules that are difficult to build and optimize. Recently, Transfuser~\cite{prakash2021multi} utilizes Transformer to encode the global relationships of point cloud and image features, whose scalability is limited by the high computational cost.

\noindent\textbf{Attention mechanisms.}
Attention is defined as the weighted sum of features at multiple positions. Attention mechanisms such as SENet ~\cite{hu2018squeeze}, CBAM~\cite{woo2018cbam}, and non-local operation~\cite{wang2018non} that exploit channel-wise and spatial attention to capture long-range dependencies have become useful plug-ins. Recently, attention-based architectures of stacked self-attention and cross-attention modules like Transformer~\cite{vaswani2017attention}, ViT~\cite{vit}, and DETR~\cite{carion2020end} have been proposed to draw the global dependencies and shown great potential in 2D~\cite{deit,setr} and 3D tasks~\cite{engel2020point,zhao2020point,pan20213d,guo2021pct} including classification, detection, and segmentation. However, their memory and computation load boost at vast key element numbers and thus hamper the model scalability. Therefore, variants emerge to solve these problems~\cite{xformer}, such as local-attention variants as Swin Transformer~\cite{liu2021Swin} and Deformable DETR~\cite{zhu2020deformable} and low-rank methods as Performer~\cite{choromanski2020rethinking} and Linear Transformer~\cite{katharopoulos2020transformers}.

\begin{figure*}[th!]
    \centering
    \includegraphics[width=12cm]{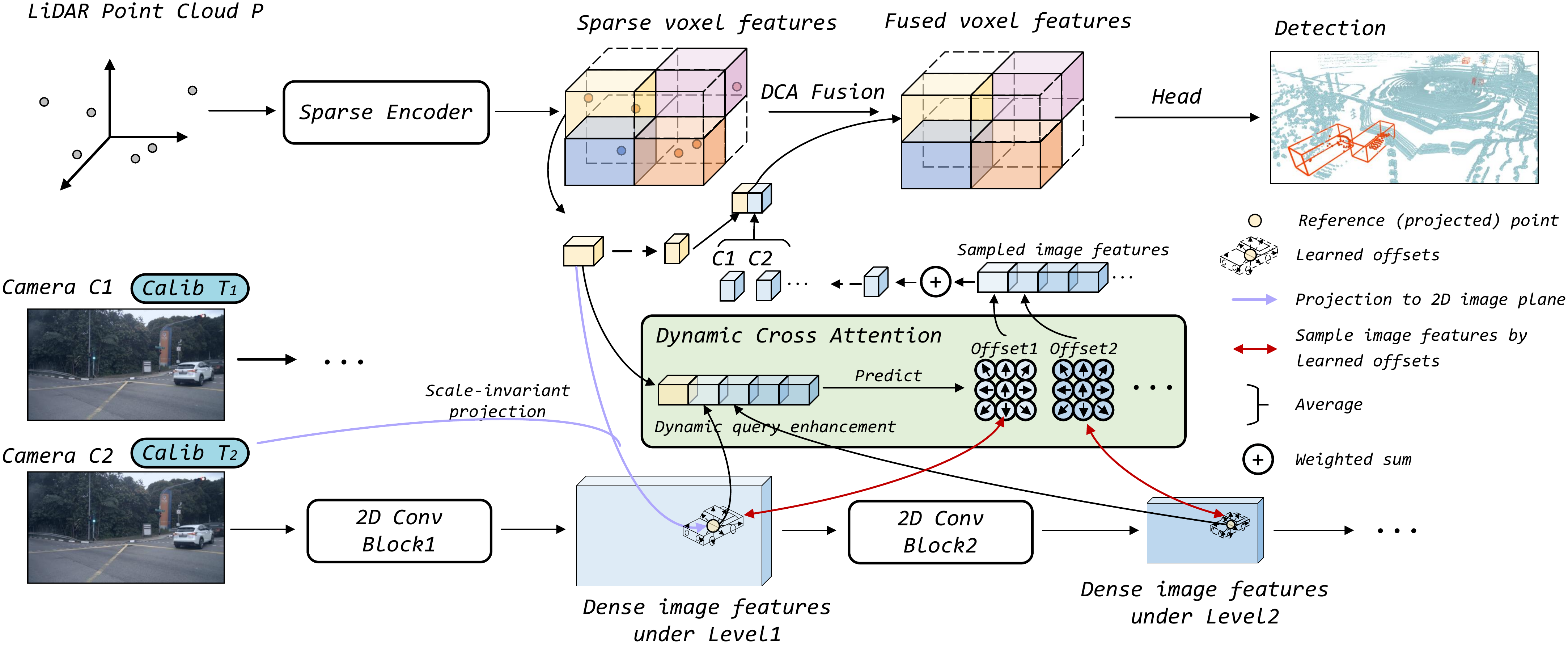}
    \caption{The DCAN architecture and DCA module. The DCAN architecture consists of an image branch for image feature extraction, a LiDAR branch as the 3D detector, and a DCA module for fusion. In the DCA module, multi-level image features are fused into the sparse 3D feature. 1) Each 3D feature is aligned with a reference point on the 2D plane and initial image features by the scale-invariant projection. 2) Then the \textit{dynamic query enhancement} incorporates the initial feature pairs to generate a query feature that is used to predict offsets towards the neighborhood and adjacent feature levels. 3) Finally, the image features on the predicted offset locations are sampled and fused with the original 3D feature, the result of which is fed into the following 3D detector.}
    \label{fig:network}
\end{figure*}


\section{Dynamic Cross Attention Network}

\subsection{Overview}
As shown in Fig.~\ref{fig:network} , the proposed Dynamic Cross Attention Network (DCAN) consists of an image branch for image feature extraction, a LiDAR branch as the 3D detector, and a Dynamic Cross Attention (DCA) module for fusion. DCAN first extracts image and LiDAR features as described in Sec.~\ref{feature_extaction}. Then our \textit{one-to-many} DCA module fuses the extracted multi-level image features into the LiDAR feature, whose output is seamlessly fed into the 3D detector for the final training and also inference. The details are described in Sec.~\ref{DCA}. Furthermore, network details including DCAN's flexible backbones that enable fusion as a plug-in and the joint-training strategy are described in Sec.~\ref{network_detail}.

\subsection{Multi-modality Feature Extraction}
\label{feature_extaction}
DCAN takes as inputs a point cloud with a set of RGB images that are captured by $K$ cameras and whose projection matrices obtained by extrinsic and intrinsic calibration are known. The network starts from feature extraction. 

\noindent\textbf{Multi-level image features.} The classic CNN architectures for image feature extraction such as VGG~\cite{vgg} and ResNet~\cite{resnet} generate feature pyramids from low to high levels. The low-level features contain spatial details while the high-level features provide strong semantics. Among existing fusion methods, MVXNet~\cite{sindagi2019mvx} chooses high-level semantics in VGG16 while PointAugmenting~\cite{wang2021pointaugmenting} adopts the shallow layer of CenterNet~\cite{duan2019centernet} for fine-grained details. To take advantage of both low-level details and high-level semantics, we use multi-level image features from ResNet50 for fusion. We denote the four levels of image features from ResNet50 as $\mathcal{I}=\{\mathrm{I}_1^k, \mathrm{I}_2^k, \mathrm{I}_3^k, \mathrm{I}_4^k\} \subset \mathbb{R}^{H \times W \times C}$, whose strides are 4, 8, 16, 32 respectively. $k$ denotes the camera. 

\noindent\textbf{LiDAR feature extraction.} The general 3D architecture consists of stacked building blocks. Within each block, point-based methods extract and aggregate point-wise features~\cite{qi2017pointnet} while voxel-based methods conduct sparse convolutions~\cite{liu2015sparse}. Due to its geometric nature, 3D LiDAR features are associated with spatial coordinates under all kinds of representations including points, voxels, and grids. 
We define a 3D feature from a 3D detector with $N$ voxels, points or grids as $\mathrm{R}=\{\mathrm{F} \in {\mathbb{R}^{{N} \times C}}, \mathrm{P} \in {\mathbb{R}^{{N} \times 3}} \}$, where $\mathrm{F} $ denotes the features and $\mathrm{P}$ denotes the associated coordinates. 

\subsection{Dynamic Cross Attention Module}
\label{DCA}


As in Fig.~\ref{fig:network} , the Dynamic Cross Attention (DCA) module is proposed to fuse the extracted cross-modality features. Each 3D feature is aligned with a reference point on the image plane by the scale-invariant projection that is consistent on different scales of image features. Then for each reference point, the \textit{dynamic query enhancement} (DQE) generates a query feature that is used to predict multiple offsets and weights towards the neighborhood on multi-level image features. With the learned offsets and weights, we sample the image features and finally fuse with the original 3D feature, the result of which is fed to the following 3D detector. 

\noindent\textbf{Scale-invariant projection.} To align each 3D feature with 2D points on multi-level image feature maps, we conduct scale-invariant projection based on the calibration. The projected 2D points, \textit{i.e.} reference points, serve as the initial clue for the following fusion. Define the calibration matrices for $K$ cameras as $\mathcal{T} = \{T^1, \ldots, T^K\} \subset \mathbb{R}^{4 \times 3}$. We project the points onto the 2D planes by 
\begin{equation}
    \label{eq:pro}
    \overline{\mathrm{P}} = \mathrm{P} \oplus \mathbbm{1}, \overline{\mathrm{P}}^k_\mathrm{ref} = T^k\overline{\mathrm{P}},
\end{equation}
where $\oplus$ denotes concatenate and $\overline{\mathrm{P}}$ is the homogeneous counterpart of $\mathrm{P}$. Then $\overline{\mathrm{P}}^k_\mathrm{ref} \in \mathbb{R}^{{N} \times 3}$ can be normalized by its third dim, which becomes the homogeneous coordinates of $\mathrm{P}^k_\mathrm{ref} \in  \mathbb{R}^{{N} \times 2}$, where $\overline{\mathrm{P}}^k_\mathrm{ref} = \mathrm{P}^k_\mathrm{ref} \oplus \mathbbm{1}$ . The coordinates of each $\mathbf{p}_{\mathrm{ref}}^{kn} \in \mathrm{P}^k_\mathrm{ref}$ are normalized to range [0, 1] by dividing the image size, and thus become invariant on different strides of image feature maps.

\noindent\textbf{Dynamic query enhancement.} 
Since the 3D feature and the initial multi-scale image features on the reference point both correspond to that reference point, we incorporate these features to generate a query feature for the following offset prediction. Compared with the \textit{one-to-one} mapping that is fixed by and unaware of the calibration, our resulting query feature contains cross-modality 3D geometry and the initial image neighborhood contexts, providing comprehensive information for the offset prediction and thus enhancing DCA's tolerance to the initial misalignment.

For each reference point $\mathbf{p}_{\mathrm{ref}}^{kn}$, the associated LiDAR feature is $\boldsymbol{f}^n \in \mathrm{F}$. We fetch the image features by bilinear interpolation:
\begin{equation}
    \label{eq:bilinear}
    \boldsymbol{I}_l^{kn} = f_{bil}(\mathrm{I}_l^k, \mathbf{p}_{\mathrm{ref}}^{kn}),
\end{equation}
where $\boldsymbol{I}_l^{kn}$ is the feature for $n$-th reference point on the $l$-th level of image from $k$-th camera. For each camera view $k$, we first use $1 \times 1$ convolution and  MLP to unify the feature channels of $\boldsymbol{I}_l^{kn}$ and $\boldsymbol{f}^n$ respectively. Then we obtain an enhanced query feature by
\begin{equation}
    \label{eq:query}
    \boldsymbol{f}_{\mathrm{query}}^{kn} = \text{LN}(\text{MLP}(\boldsymbol{f}^n)) \oplus \text{LN}(\text{Conv}_{1\times 1}(\boldsymbol{I}_l^{kn})), 
    \forall l,
\end{equation}
where $\oplus$ denotes concatenate operation, and $\text{LN}$ denotes LayerNorm~\cite{LN}.  

\begin{algorithm}[tb]
\caption{Dynamic Cross Attention Module}
\label{alg:DCA}
\textbf{Input}: Sparse LiDAR feature $\mathrm{R} \leftarrow \{\mathrm{F} \in {\mathbb{R}^{{N} \times C}}, \mathrm{P} \in {\mathbb{R}^{{N} \times 3}} \}$, projected reference points $\mathrm{P}^k_\mathrm{ref}$, and multi-level image features  $\mathcal{I} \leftarrow \{\mathrm{I}_1^k, \mathrm{I}_2^k, \mathrm{I}_3^k, \mathrm{I}_4^k\} \subset \mathbb{R}^{H \times W \times C}$

\textbf{Output}: Sparse fusion feature $\bar{\mathrm{R}}$
\begin{algorithmic}[1] 
 \FOR{$n \leftarrow 1$ to $N$}
    \STATE $\boldsymbol{f}^n \in \mathrm{F}, \mathbf{p}^n \in \mathrm{P}$
    \FOR{$k \leftarrow 1$ to $K$}
        \STATE $\mathbf{p}_{\mathrm{ref}}^{kn} \leftarrow \text{Equ.~\ref{eq:pro}} \in \mathrm{P}^k_\mathrm{ref}$ \textit{// scale-invariant projection}
        \STATE $\boldsymbol{f}_{\mathrm{query}}^{kn} \leftarrow \text{Equ.~\ref{eq:query}}$ \textit{// dynamic query enhancement}
        \STATE $\Delta \mathbf{p}^{kn} \leftarrow  \text{MLP}(\boldsymbol{f}_{\mathrm{query}}^{kn}) \in \mathbb{R}^{(L\times M \times D)\times 2}$ \textit{// offsets prediction}
        \STATE $\mathbf{w}^{kn} \leftarrow  \text{MLP}(\boldsymbol{f}_{\mathrm{query}}^{kn}) \in \mathbb{R}^{L\times M \times D}$ \textit{// weights prediction}
        \STATE $\boldsymbol{I}_{\mathrm{value}}^{kn} \leftarrow \text{Equ.~\ref{eq:one-to-many}}$ \textit{// one-to-many attention}
    \ENDFOR
    \STATE $\boldsymbol{I}_{\mathrm{value}}^n \leftarrow  \text{Equ.~\ref{eq:viewmean}}$ \textit{// mean of valid views}
    \STATE $\bar{\boldsymbol{f}^n} \leftarrow \text{FFN}(\boldsymbol{f}^n + \boldsymbol{I}_{\mathrm{value}}^n) \in \bar{\mathrm{F}}$
 \ENDFOR
 \RETURN{$\bar{\mathrm{R}} \leftarrow \{\bar{\mathrm{F}}\in{\mathbb{R}^{{N} \times C}}, \mathrm{P} \in {\mathbb{R}^{{N} \times 3}} \}$}

\end{algorithmic}
\end{algorithm}

\noindent\textbf{Dynamic cross attention.} 
Then $\boldsymbol{f}^{kn}_{\mathrm{query}}$ is fed into two MLP layers to predict the sampling offsets $\Delta \mathbf{p}^{kn} \in \mathbb{R}^{2}$ and the corresponding attention weights $\mathbf{w}^{kn} \in \mathbb{R}$ respectively. $D$ points towards $M$ directions on $L$ levels of image features that compose a total number of $D*M*L$ offsets and weights are predicted. We sample the features on all of the learned offset locations and conduct a weighted addition of them by the learned weights:
\begin{equation}
    \label{eq:one-to-many}
    \boldsymbol{I}_{\mathrm{value}}^{kn} = \sum_{l,m,d} \mathbf{w}^{kn}_{lmd} \cdot f_{bil}(\mathrm{I}_l^k ,\mathbf{p}_{\mathrm{ref}}^{kn} + \Delta \mathbf{p}^{kn}_{lmd}), 
\end{equation}
where $\sum_{l,d}\mathbf{w}^{kn}_{lmd} = 1$, implemented by softmax. Since a 3D point can have one or two valid $\mathbf{p}_{\mathrm{ref}}^{kn}$ among all $K$ camera views considering overlap, we conduct mean operation among the $K^*$ valid reference points for each 3D point:
\begin{equation}
    \label{eq:viewmean}
    \boldsymbol{I}_{\mathrm{value}}^{n} = \frac{1}{K^*}\sum_{k^*} \boldsymbol{I}_{\mathrm{value}}^{{k^*}n}, \forall n,
\end{equation}
Finally, the dynamic cross attention is described as:
\begin{equation}
    \bar{\mathrm{F}} = \text{CrossAttn}(\mathrm{R}, \mathcal{L}) = \text{FFN}(\boldsymbol{f}^n + \boldsymbol{I}_{\mathrm{value}}^n), \forall n,
\end{equation}
where $\text{FFN}$ denotes the feed-forward layer as in Transformer~\cite{vaswani2017attention}. Note that we add $\boldsymbol{I}^n_{\mathrm{value}}$ to the raw LiDAR feature instead of the enhanced query feature. The resulting sparse fusion features are fed into the followed 3D detector for supervision and also inference. The entire DCA module is described as \textbf{Alg.~\ref{alg:DCA}}.


\subsection{Network Details}
\label{network_detail}

\textbf{Flexible backbones.}
The choice of backbones is flexible in DCAN.  For the LiDAR backbone, DCAN can adapt to multiple representations including points, grids, and voxels. A point feature as in 3DSSD~\cite{yang20203dssd} naturally associates with 3D coordinates to project the reference point. For a voxel feature as in Centerpoint~\cite{yin2021center} and SECOND~\cite{yan2018second}, the associated coordinates could be the mean of points within the voxels. For a grid feature as in PointPillar~\cite{lang2019pointpillars}, we take those pillars with raw 3D points in them as query features and take the mean of points as the reference point. 
For the image backbone, we use a 2D detection network instead of 2D segmentation networks considering the high costs of 2D segmentation annotations.

\noindent\textbf{Joint-training strategy.} In many existing fusion works, pre-trained backbones of Faster-RCNN~\cite{ren2015faster} or HTCNet~\cite{HTC} are utilized for image feature extraction while the model parameters are frozen or explicitly supervised by 3D detectors. Considering the complementary natures of point clouds and images, we adopt a joint-training strategy. An image detector works as an auxiliary loss to directly supervise the image feature extractor by the 2D bounding box annotations. Such supervision is cost-free for inference while benefiting the 3D detector in an explicit manner.


\section{Experiments}
In this section, we test our model on the standard benchmark dataset nuScenes~\cite{nuscenes2019} and KITTI~\cite{KITTI}. We compare our network with the state-of-the-art methods on the nuScenes test set. Extensive results of the proposed DCA serving as a plug-in module on point-based, grid-based, and voxel-based 3D detectors prove its generalized effectiveness. In the ablation study, we validate the effectiveness of \textit{one-to-many} mapping and DQE, and explore the selection of hyper-parameters. 
A disturbance during training and inference shows our \textit{one-to-many} DCA's robustness against the calibration error compared with the \textit{one-to-one} mapping.

\subsection{Dataset and Metrics}

\textbf{nuScenes} is a challenging large-scale dataset developed for autonomous driving. nuScenes consists of 1000 scenes of 20s duration, 850 of which for training and validation and 150 for testing. The point clouds are annotated with 3D bounding boxes and semantic labels every 5 frames. Each annotated frame is associated with images from six cameras [\textit{front, front\_right, front\_left, back, back\_right, back\_left}] by the extrinsic and intrinsic matrices. 10 out of 23 annotation classes are used for evaluation. Except for 3D annotations, nuScenes also released a 2D dataset named nuImages. It contains 93,000 images with 2D bounding boxes of classes consistent with 3D detection.

\noindent\textbf{KITTI} is a relatively small dataset that offers synced LiDAR points and front-view images. KITTI provides 3D bounding boxes of cars, pedestrians, and cyclists. Only objects visible in the image are annotated. In our experiments, we follow the practice of MV3D~\cite{MV3D} and SECOND~\cite{yan2018second} to split the 7481 training samples into a training set of 3712 samples and a validation set of 3769 samples.

\noindent\textbf{Evaluation metrics.} We follow the official metrics given by nuScenes and KITTI. nuScenes evaluates Average Precision(AP) and five consolidated error types Translation Error(ATE), Average Scale Error (ASE), Average Orientation Error (AOE), Average Velocity Error (AVE), and Average Attribute Error (AAE), all of which are unified to a scalar score, nuScenes detection score (NDS)~\cite{nuscenes2019}. On KITTI, we evaluate the mAP of bird-eye view (BEV5) with 0.7 IOU at easy, moderate, and hard difficulties. 

\begin{table*}[ht!]
    \centering
    \renewcommand\arraystretch{1.2}
\resizebox{\textwidth}{!}{

\begin{tabular}{ l| c| c| c| c c c c c c c c c c}

    \hline
\multicolumn{1}{c|}{\multirow{2}{*}{Methods}} & \multirow{2}{*}{Modality} & \multirow{2}{*}{NDS(\%)} & \multirow{2}{*}{mAP(\%)} & \multicolumn{10}{c}{AP (\%)}                                                                                                                                                                                                                                                     \\
\multicolumn{1}{c|}{}                         &                           &                          &                          & \multicolumn{1}{l}{Car} & \multicolumn{1}{l}{Truck} & \multicolumn{1}{l}{Bus} & \multicolumn{1}{l}{Trailer} & \multicolumn{1}{l}{C.V.} & \multicolumn{1}{l}{Ped.} & \multicolumn{1}{l}{Moto.} & \multicolumn{1}{l}{Bicycle} & \multicolumn{1}{l}{T.C.} & \multicolumn{1}{l}{Barrier} \\ \hline
PointPillars~\cite{lang2019pointpillars}                                  & L                         & 55.0                     & 40.1                     & 76.0                    & 31.0                      & 32.1                    & 36.6                        & 11.3                     & 64.0                    & 34.2                     & 14.0                        & 45.6                    & 56.4                        \\
3DSSD~\cite{yang20203dssd}                                         & L                         & 56.4                     & 46.2                     & 81.2                    & 47.2                      & 61.4                    & 30.5                        & 12.6                     & 70.2                    & 36.0                     & 8.6                         & 31.1                    & 47.9                        \\
CBGS~\cite{cbgs}                                          & L                         & 63.3                     & 52.8                     & 81.1                    & 48.5                      & 54.9                    & 42.9                        & 10.5                     & 80.1                    & 51.5                     & 22.3                        & 70.9                    & 65.7                        \\
HotSpotNet~\cite{hotspot}                                    & L                         & 66.0                     & 59.3                     & 83.1                    & 50.9                      & 56.4                    & 53.3                        & 23.0                     & 81.3                    & 63.5                     & 36.6                        & 73.0                    & 71.6                        \\
Centerpoint~\cite{yin2021center}                                   & L                         & 67.3                     & 60.3                     & 85.2                    & 53.5                      & 63.6                    & 56.0                        & 20.0                     & 54.6                    & 59.5                     & 30.7                        & 78.4                    & 71.1                        \\ \hline
PointPainting~\cite{vora2020pointpainting}                                 & L + C                    & 58.1                     & 46.4                     & 77.9                    & 35.8                      & 36.2                    & 37.3                        & 15.8                     & 73.3                    & 41.5                     & 24.1                        & 62.4                    & 60.2                        \\
3DCVF~\cite{yoo20203d}                                         & L + C                    & 62.3                     & 52.7                     & 83.0                    & 45.0                      & 48.8                    & 49.6                        & 15.9                     & 74.2                    & 51.2                     & 30.4                        & 62.9                    & 65.9                        \\
MOCA~\cite{zhang2020exploring}                                          & L + C                    & 70.9                     & 66.6                     & 86.7                    & \textbf{58.6}                      & \textbf{67.2}                    & 60.3                        & \textbf{32.6}            & 87.1                    & 67.8                     & 52                          & 81.3                    & 72.3                        \\
PointAugmenting~\cite{wang2021pointaugmenting}                               & L + C                    & 71.1                     & 66.8                     & 87.5                    & 57.3                      & 65.2                    & \textbf{60.7}                        & 28.0                     & 87.9                    & \textbf{74.3}                     & 50.9                        & 83.6                    & \textbf{72.6}               \\ \hline
DCAN(ours)                                   & L + C                    & \textbf{71.6}            & \textbf{67.3}                    & \textbf{87.8}           & 58.5                      & 66.8          & 60.5                        & 22.5                     & \textbf{88.8}           & 73.2                     & \textbf{53.0}                    & \textbf{84.4}                    & 71.1                        \\ \hline
    \end{tabular}
}
\caption{Performance comparisons on nuScenes detection test set. "L" and "C" denote the input modality of LiDAR and cameras respectively. We report NDS, mAP, and mAP for each class. Abbreviations: construction vehicle (C.V.), pedestrian (Ped.), motorcycle (Moto.), and traffic cone (T.C.).}
\label{tab:nus_det}
\end{table*}

\subsection{Implementation Details and Results} \label{sec:res_eval}
\textbf{Model details.} Our model consists of a ResNet50 to extract image features, a flexible 3D detector, and a DCA module. We use ResNet50 pre-trained by HTCNet on nuImages for nuScenes, and for KITTI we use ResNet50 pre-trained by Faster R-CNN~\cite{ren2015faster} as in MOCA~\cite{zhang2020exploring}. In the DCA module, we have three hyper-parameters $L$, $M$, and $D$ that denote the levels of the image features for fusion, the number of directions, and the number of offsets towards each direction on each feature level, respectively. We use all levels for fusion, \textit{i.e.} $L=4$, and set $M$ and $D$ as 8 and 4 respectively. More details are in the ablation study. In the joint training strategy, we choose FCOS~\cite{tian2019fcos} as our 2D detector. 

\noindent\textbf{Comparison to state-of-the-arts} As in \textbf{Tab.~\ref{tab:nus_det}}, we compare DCAN with the state-of-the-art methods on the nuScenes test set. We adopt Centerpoint~\cite{yin2021center} as the 3D detector and plug the DCA module into the 2\textit{nd} block for test submission. We set the point cloud range as $[-54m, 54m]$ for the $X$ and $Y$ axis, and $[-5m, 3m ]$ for the $Z$ axis. The voxel size is set $[0.075m, 0.075m, 0.2m]$. We use 10 sweeps for point cloud enhancement. We adopt a group balancing strategy called CBGS~\cite{cbgs}. To save memory usage, the input images are resized from $(1600, 900)$ to $(800, 450)$. We also use data augmentation: 3D random rotation of $[-22.5^\circ, 22.5^\circ]$, 3D random scale of $[0.95, 1.05]$, 2D and 3D random flip. We adopt hybrid optimizers following MOCA~\cite{zhang2020exploring}. An AdamW~\cite{adamw} with the initial learning rate of $1e^{-4}$ and a weight decay of $0.01$ is used to optimize the DCA and the 3D detector, while an SGD~\cite{sgd} with a learning rate of 0.002, a momentum of 0.9, and a weight decay of $1e^{-4}$ is used to optimize the 2D detector including ResNet50. We train the model on 8 GeForce RTX 3090 GPUs at a batch size of 2 for 20 epochs with mixed precision training~\cite{micikevicius2017mixed}. 

Before this paper is submitted, our method outperforms published methods for the nuScenes detection challenge. 
We \textbf{do not} use GT-Aug or model ensemble. As in Tab.~\ref{tab:nus_det}, DCAN's performance on the small objects including pedestrian, bicycle, and traffic cone outperforms the second-best methods for 0.9, 2.1, and 0.8 respectively. Considering that bicycles only count for 1.02\% of all the annotated objects and that MOCA~\cite{zhang2020exploring} and PointAugmenting~\cite{wang2021pointaugmenting} both adopt GT-Aug to alleviate the class imbalance problem, DCAN works outstandingly for bicycles. DCAN models the dynamic alignment of point clouds and images, which is caused by the dynamic optimization of calibration. Therefore, compared with the point-wise methods whose \textit{one-to-one} mapping is fixed by the given calibration, DCAN performs especially well for small objects that are relatively sensitive to the calibration error. 

\noindent\textbf{DCA as a plug-in.} We add DCA to LiDAR-only detectors including grid-based, voxel-based, and point-based methods on nuScenes and KITTI dataset. The detectors include PointPillars~\cite{lang2019pointpillars}, Centerpoint~\cite{yin2021center}, SECOND~\cite{yan2018second}, and 3DSSD~\cite{yang20203dssd}. The results are shown in \textbf{Tab.~\ref{tab:nus_cmp}} and \textbf{Tab.~\ref{tab:kitti_cmp}}. The reports of baselines result from the implementation by MMDetection3D~\cite{mmdet3d2020}, and we remove the GT-Aug from the data augmentation for a fair comparison. We train the model for the same training epochs with and without DCA. We only compare Car results on KITTI because 3DSSD only reports Car results. We report a complete comparison with SECOND on KITTI in the supplementary.

\begin{table*}[h!]
\centering
\resizebox{\textwidth}{!}{%
\begin{tabular}{l|cc|cccccccccc}
\hline
\multicolumn{1}{c|}{Methods} & NDS & mAP & Car & Truck & Bus & Trailer & C.V. & Ped. & Moto. & Bicycle & T.C & Barrier \\ \hline \hline
PointPillars & 53.3 & 40.0 & 81.4 & 31.8 & 1.3 & 44.4 & 16.6 & 52.8 & 39.6 & 9.7 & 71.4 & 30.8 \\
PointPillars+DCA & 63.3 & 56.7 & 87.8 & 44.9 & 6.4 & 47.3 & 28.9 & 73.8 & 64.7 & 46.9 & 85.7 & 66.3 \\\hline
Gains of fusion & {\color[HTML]{FE0000} +10.0} & {\color[HTML]{FE0000} +16.7} & {\color[HTML]{FE0000} +6.3} & {\color[HTML]{FE0000} +13.1} & {\color[HTML]{FE0000} +5.1} & {\color[HTML]{FE0000} +2.9} & {\color[HTML]{FE0000} +12.3} & {\color[HTML]{FE0000} +21.0} & {\color[HTML]{FE0000} +25.1} & {\color[HTML]{FE0000} +37.2} & {\color[HTML]{FE0000} +14.3} & {\color[HTML]{FE0000} +35.5} \\ \hline \hline
Centerpoint & 61.4 & 54.1 & 82.7 & 52.5 & 11.3 & 62.0 & 25.7 & 57.7 & 59.7 & 42.4 & 79.5 & 58.1 \\
Centerpoint+DCA & 67.0 & 64.1 & 86.6 & 59.6 & 15.0 & 66.5 & 35.2 & 68.2 & 75.2 & 65.6 & 84.7 & 71.3 \\ \hline
Gains of fusion & {\color[HTML]{FE0000} +5.6} & {\color[HTML]{FE0000} +10.0} & {\color[HTML]{FE0000} +3.9} & {\color[HTML]{FE0000} +7.1} & {\color[HTML]{FE0000} +3.7} & {\color[HTML]{FE0000} +4.5} & {\color[HTML]{FE0000} +9.5} & {\color[HTML]{FE0000} +10.5} & {\color[HTML]{FE0000} +15.5} & {\color[HTML]{FE0000} +23.2} & {\color[HTML]{FE0000} +5.2} & {\color[HTML]{FE0000} +13.2} \\ \hline
\end{tabular}%
}
\caption{Comparisons of results on nuScenes validation set. We report NDS, mAP, and mAP for each class.}
\label{tab:nus_cmp}
\end{table*}

\begin{table}[h!]
\centering
\setlength{\tabcolsep}{6mm}{
\begin{tabular}{c|ccc}
\hline
\multirow{2}{*}{Methods} & \multicolumn{3}{c}{$\text{AP}_{\text{BEV}}(\text{IOU}=0.7)$} \\ \cline{2-4} 
                         & Easy   & Mod.   & Hard  \\ \hline \hline
SECOND~\cite{yan2018second}&	90.2&	86.5&	79.9\\ 
SECOND+DCA&	90.4&	88.2&	87.0\\ \hline
Gains of fusion&	{\color[HTML]{FE0000} +0.2}&	{\color[HTML]{FE0000} +1.7}&	{\color[HTML]{FE0000} +7.1}\\ \hline \hline
3DSSD~\cite{yang20203dssd}&	89.6&	86.1&	80.6\\ 
3DSSD+DCA&	90.5&	87.3&	80.7\\ \hline
Gains of fusion&	{\color[HTML]{FE0000} +0.9}&	{\color[HTML]{FE0000} +1.2}&	{\color[HTML]{FE0000} +0.1}\\ \hline
\end{tabular}}
\caption{Comparisons of Car Results on KITTI validation set.}
\label{tab:kitti_cmp}
\end{table}

In Tab.~\ref{tab:nus_cmp}, DCA boosts the NDS of 10.0 for PointPillars and 5.6 for Centerpoint. While the detection performance of the DCA module generalizes well on every class, the small objects are improved especially great with the gains mostly over 10 AP. In general, the classes that originally perform worse and have fewer samples receive relatively greater gains. The largest increase for both methods is on the bicycle class, which achieves 37.2 and 23.2, respectively. In Tab.~\ref{tab:kitti_cmp}, while limited by the small size of KITTI, DCA still raises the performance of both SECOND and 3DSSD. The more difficult Moderate and Hard objects receive greater gains. The general improvements and especial benefits for small objects on the 3D detectors of all types of 
representations manifest the generalization of DCA and its great potential to serve as a plug-in fusion module. 

\noindent\textbf{Latency.} Since DCA is designed as a plug-in module, the latency of the whole network mainly relies on the chosen image and LiDAR backbones,
as shown in \textbf{Tab.~\ref{albation_efficiency}}. 
The \#params of PointPillar+DCA and Centerpoint+DCA are 348M and 241M, respectively. The DCA modules cost 12.7\% and 10.9\% of overall respectively. The efficiency of DCA can be improved by processing the projection and bilinear sampling of six views as a batch, which enables sampling to cost under 2 \textit{ms} in our experiments. Mixed precision training~\cite{micikevicius2017mixed} saves 10.5\% memory and 6.7\% training time while hurts 1.3 NDS for Centerpoint+DCA. 
\begin{table}[h!]
\centering
\setlength{\tabcolsep}{2mm}{
\begin{tabular}{c|c|cc|cc}
\hline
Method & Overall  & Image &  LiDAR & \#voxels & DCA \\ \hline  \hline
PointPillar+DCA & 119.3 \textit{ms}   & 53.8 \textit{ms}  & 50.4 \textit{ms} &  20,000 & 15.1 \textit{ms} \\ 
Centerpoint+DCA & 167.6 \textit{ms}  & 53.8 \textit{ms}  & 95.5 \textit{ms} & 40,000  & 18.3 \textit{ms} \\ \hline
\end{tabular}}
\caption{Latency of DCA fusion and other modules on a 2080Ti.}
\label{albation_efficiency}
\end{table}

\subsection{Ablation Study}
\label{sec:ablation}
In this section, we conduct a detailed ablation study of each DCA component and validate our DCA module's robustness against the calibration error. We adopt ResNet50 for image feature extraction and Centerpoint as the 3D detector. The point cloud range is set as $[-51.2m, 51.2m]$ for the $X$ and $Y$ axis, and $[-5m, 3m ]$ for the $Z$ axis, and the voxel size is set as $[0.1m, 0.1m, 0.2m]$ throughout the ablation study. 

\begin{table}[h!]
\centering
\renewcommand\arraystretch{1.1}
\begin{tabular}{l|c|c|c|c|cc}
\hline
Ablation factor      & Offset learning                        & Collection of $L$               & $D$                & $M$                & NDS  & mAP  \\ \hline
\multirow{2}{*}{\textit{\begin{tabular}[c]{@{}l@{}}one-to-one mapping\\ w/ or w/o offset learning\end{tabular}}}   & \xmark                  & \multirow{2}{*}{\{1\}}          & \multirow{2}{*}{1} & \multirow{2}{*}{1} & 64.5 & 59.2 \\
     & \cmark                  &                                 &                    &        & 65.1     &   61.6   \\ \hline \hline
\multirow{4}{*}{\textit{\begin{tabular}[c]{@{}l@{}}image feature levels \\ used for fusion\end{tabular}}}                                        & \multirow{4}{*}{\cmark} & \{1\}                           & \multirow{4}{*}{8} & \multirow{4}{*}{4} & 65.9 & 62.7 \\
                                                                                                                                                 &                                        & \{1, 2\}                        &                    &        &   67.2  &  64.8    \\
                                                                                                                                                 &                                        & \{1, 2, 3\}                     &                    &         & 67.8  &   65.0    \\
                                                                                                                                                 &                                        & \{1, 2, 3, 4\}                  &                    &                    & 68.0   & 64.9 \\ \hline \hline
\multirow{3}{*}{\textit{the number of directions}}                                                                                               & \multirow{3}{*}{\cmark} & \multirow{3}{*}{\{1, 2, 3, 4\}} & 2                  & \multirow{3}{*}{4} & 67.5 & 64.4 \\
                                                                                                                                                 &                                        &                                 & 4                  &                    & 67.7 & 64.8 \\
                                                                                                                                                 &                                        &                                 & 8                  &                    & 68.0   & 64.9 \\ \hline \hline
\multirow{3}{*}{\textit{\begin{tabular}[c]{@{}l@{}}the number of offsets towards \\ each direction on each image \\ feature level\end{tabular}}} & \multirow{3}{*}{\cmark} & \multirow{3}{*}{\{1, 2, 3, 4\}} & \multirow{3}{*}{8} & 4                  & 68.0   & 64.9 \\
                                                                                                                                                 &                                        &                                 &                    & 8                  & 67.9 & 64.9 \\
                                                                                                                                                 &                                        &                                 &                    & 16                 & 67.6 & 64.4 \\ \hline
\end{tabular}
\caption{Ablation study of \textit{one-to-many} mapping. }
\label{tab:ablation1}
\end{table}

\noindent\textbf{One-to-many mapping.} In \textbf{Tab. ~\ref{tab:ablation1}}, we conduct experiments about the offset learning and the hyper-parameters that decide the number of learned offsets. The first row is the result of the \textit{one-to-one} mapping, for which we naively concatenate each 3D feature and its aligned image feature of stride 4, \textit{i.e.} the 1st image level, based on the calibration for fusion, following the state-of-the-art PointAugmenting~\cite{wang2021pointaugmenting}. As Tab. ~\ref{tab:ablation1} shows: 1) Learning offsets improves the \textit{one-to-one} mapping by 0.6 NDS and 2.4 mAP, which shows the necessity to optimize the initial alignment by the calibration. 2) By fusing more and all 4 levels of image features, DCAN achieves better and the best performance, respectively. 3) The performance of different $D$s and $M$s are very close where 8 and 4 are the optimal choices, respectively. Notably, thanks to our scale-invariant projection and processing the sampling and bilinear interpolation of six views as a batch, the time cost among all groups of selected $L$, $D$, and $M$ varies within \textbf{4 \textit{ms}}. The results throughout Tab. ~\ref{tab:ablation1} are with DQE.

\noindent\textbf{Dynamic query enhancement.} In Equ.~\ref{eq:query}, we can choose to use LiDAR, image, or both features as the query feature for offset and weight prediction. \textbf{Tab. ~\ref{tab:ablation2}} shows the benefits of incorporating both features. We set collection of $L$ as \{1, 2, 3, 4\}, $D = 8$, and $M = 4$ for the experiments. 
\begin{table}[h!]
\centering
\setlength{\tabcolsep}{5mm}{
\begin{tabular}{c|cc}
\hline
query feature & NDS  & mAP  \\ \hline \hline
L            & 67.4 & 64.5 \\ \hline
I            & 67.1 & 64.4 \\ \hline
L+I          & 68   & 64.9 \\ \hline
\end{tabular}}
\caption{Ablation study of the type of the query feature used for offset and weight prediction. L and I stand for LiDAR and image respectively.}
\vspace{-10mm}
\label{tab:ablation2}
\end{table}

\begin{figure}
    \centering
    \includegraphics[width=\textwidth]{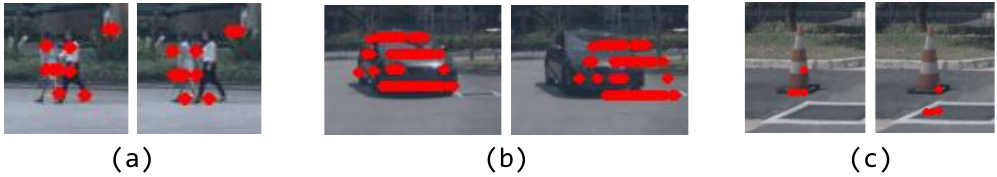}
    \caption{The left and right pictures are the projected reference points before and after disturbance, respectively.}
    \label{fig:disturbance}
\end{figure}

\noindent\textbf{Calibration error test.} As in Sec.~\ref{DCA}, we propose \textit{one-to-many} mapping to model the dynamic calibration-based alignment of point clouds and images. Besides, the DQE is proposed to provide cross-modality guidance for the offset prediction. We test our \textit{one-to-many} DCA module without and with DQE, and compare them with the \textit{one-to-one} mapping. In the case of DCA without DQE, we simply use the original 3D feature for the offset prediction. To implement the \textit{one-to-one} mapping, we concatenate each 3D feature and its aligned image feature of stride 4 as described in Tab. ~\ref{tab:ablation1}. Besides, to test the robustness against the calibration error, we introduce disturbance to the calibration during training and validation. 50\% possibility of disturbance that obeys an even distribution within $2^\circ$ rotation and $20cm$ translation are added to the initial calibration matrices. An example is given by Fig.~\ref{fig:disturbance}. 

As is shown in \textbf{Tab.~\ref{tab:offset_ablation}}: 1) With and without validation disturbance, the \textit{one-to-many} models without DQE have drop of 0.1 and 0.6 NDS, respectively, which are 10 times and 1.8 times lower than the drop of 1.8 and 1.7 NDS of the \textit{one-to-one} model, respectively. Such comparisons show the tolerance to the calibration error of the \textit{one-to-many} mapping. 2) With and without validation disturbance, the DQE further gains 0.3 and 0.4 NDS based on the \textit{one-to-many} mapping, which proves DQE's ability to further strengthen the robustness of the DCA module. 3) Both the proposed models have lower losses of disturbance on most of the classes compared with the \textit{one-to-one} model. In summary, it is safe to say that DCA develops robustness against the calibration error compared with the \textit{one-to-one} methods. 

\begin{table}[ht!]
    \Large
    \centering
    \renewcommand\arraystretch{1.2}
\begin{subtable}[h]{\textwidth}
\resizebox{\textwidth}{!}{
\begin{tabular}{c|cc|cc|cccccccccc}
\hline
Ablation Study of DCA & \multicolumn{1}{c|}{Training Disturbance} & Validation Disturbance & NDS & mAP & Car & Truck & Bus & Trailer & C.V. & Ped. & Moto. & Bicycle & T.C & Barrier \\ \hline  \hline
 & \xmark & \xmark & 64.5 & 59.2 & 83.4 & 53.1 & 64.8 & 28.3 & 12.3 & 82.6 & 66.6 & 52.7 & 68.9 & 66.3 \\ \cline{2-15} 
 & \cmark & \xmark & 62.7 & 56.3 & 82.5 & 52.5 & 62.7 & 28.7 & 11.0 & 80.7 & 59.9 & 46.8 & 61.5 & 62.7 \\ \cline{2-15} 
\multirow{-3}{*}{\textit{one-to-one}} & \multicolumn{2}{c|}{loss of disturbance} & -1.8 & -2.9 & -0.9 & -0.6 & -2.1 & +0.4 & -1.3 & -1.8 & -6.7 & -5.9 & -7.4 & -3.6 \\ \hline \hline
 & \xmark & \xmark & 67.4 & 64.5 & 86.1 & 59.0 & 67.7 & 36.4 & 17.2 & 85.1 & 75.5 & 65.0 & 72.8 & 67.7 \\ \cline{2-15} 
 & \cmark & \xmark & 67.3 & 64.3 & 86.7 & 58.9 & 70.7 & 29.9 & 14.9 & 85.0 & 73.6 & 66.1 & 72.1 & 70.2 \\ \cline{2-15} 
\multirow{-3}{*}{\textit{one-to-many}} & \multicolumn{2}{c|}{loss of disturbance} & -0.1 & -0.2 & +0.6 & -0.2 & {\color[HTML]{FE0000} +3.0} & -6.5 & -2.4 & -0.2 & -1.9 & +1.1 & {\color[HTML]{FE0000} -0.7} & {\color[HTML]{FE0000} +2.5} \\ \hline \hline
\textit{one-to-many} & \xmark & \xmark & 68.0 & 64.9 & 87.0 & 60.2 & 70.8 & 32.5 & 17.4 & 85.6 & 73.6 & 65.1 & 74.1 & 69.4 \\ \cline{2-15} 
w/ & \cmark & \xmark & 68.2 & 65.4 & 86.8 & 61.4 & 69.9 & 34.0 & 17.7 & 85.5 & 76.3 & 66.8 & 72.7 & 69.2 \\ \cline{2-15} 
DQE & \multicolumn{2}{c|}{loss of disturbance} & {\color[HTML]{FE0000} +0.2} & {\color[HTML]{FE0000} +0.5} & {\color[HTML]{FE0000} -0.2} & {\color[HTML]{FE0000} +1.2} & -0.9 & {\color[HTML]{FE0000} +1.6} & {\color[HTML]{FE0000} +0.3} & {\color[HTML]{FE0000} -0.1} & {\color[HTML]{FE0000} +2.7} & {\color[HTML]{FE0000} +1.8} & -1.3 & -0.2 \\ \hline
\end{tabular}
}
\end{subtable}

\bigskip

\begin{subtable}[h]{\textwidth}
\resizebox{\textwidth}{!}{
\begin{tabular}{c|cc|cc|cccccccccc}
\hline
Ablation Study of DCA & \multicolumn{1}{c|}{Training Disturbance} & Validation Disturbance & NDS & mAP & Car & Truck & Bus & Trailer & C.V. & Ped. & Moto. & Bicycle & T.C & Barrier \\ \hline \hline
 & \xmark & \xmark &    64.5&   59.2&   83.4&   53.1&   64.8&   28.3&   12.3&   82.6&   66.6&   52.7&   68.9&   66.3  \\ \cline{2-15} 
 & \cmark & \cmark &       62.8&   56.4&   82.5&   52.7&   63.1&   29.9&   11.0&   80.5&   59.7&   46.8&   61.9&   63.1 \\ \cline{2-15} 
\multirow{-3}{*}{\textit{one-to-one}} & \multicolumn{2}{c|}{loss of disturbance}&        -1.7&   -2.8&   -1.0&   -0.4&   -1.7&   {\color[HTML]{FE0000} +1.5}&   -1.2&   -2.0&   -7.0&   -5.9&   -7.0&   -3.2 \\ \hline \hline
 & \xmark & \xmark &        67.4&   64.5&   86.1&   59.0&   67.7&   36.4&   17.2&   85.1&   75.5&   65.0&   72.8&   67.7 \\ \cline{2-15} 
 & \cmark & \cmark &   66.8&   63.5&   85.9&   58.5&   69.6&   28.9&   14.7&   84.6&   71.6&   65.3&   71.2&   69.1 \\ \cline{2-15} 
\multirow{-3}{*}{\textit{one-to-many}} & \multicolumn{2}{c|}{loss of disturbance} &        -0.6&   -1.0&   {\color[HTML]{FE0000} -0.2}&   -0.6&   {\color[HTML]{FE0000} +1.9}&   -7.4&   -2.5&   -0.6&   -3.9&   +0.4&   {\color[HTML]{FE0000} -1.6}&   {\color[HTML]{FE0000} +1.4} \\ \hline \hline
\textit{one-to-many} & \xmark & \xmark&  68.0&   64.9&   87.0&   60.2&   70.8&   32.5&   17.4&   85.6&   73.6&   65.1&   74.1&   69.4 \\ \cline{2-15} 
w/ & \cmark & \cmark&     67.7&   64.7&   86.0&   60.9&   68.6&   33.5&   17.4&   85.1&   74.9&   65.6&   71.7&   68.5 \\ \cline{2-15} 
DQE & \multicolumn{2}{c|}{loss of disturbance} &        {\color[HTML]{FE0000} -0.2}&   {\color[HTML]{FE0000} -0.2}&   -0.9&   {\color[HTML]{FE0000} +0.7}&   -2.1&   +1.1&   {\color[HTML]{FE0000} +0.1}&   {\color[HTML]{FE0000} -0.5}&   {\color[HTML]{FE0000} +1.3} &   {\color[HTML]{FE0000} +0.6} &   -2.4&   -1.0 \\ \hline
\end{tabular}
}
\end{subtable}

\caption{Ablation study of \textit{one-to-one}, \textit{one-to-many}, and \textit{dynamic query enhancement} models under calibration error test. The smallest losses of disturbance among the three models are marked red.}
\label{tab:offset_ablation}
\vspace{-10mm}
\end{table}

\section{Limitations}
As in Sec.~\ref{sec:ablation}, we prove DCA's robustness against an extent of calibration error. However, the proposed method has the following limitations: 1) Without disturbance in training and with that in validation, while the \textit{one-to-many} models with and without DQE still have over 3 times lower drop of NDS than the \textit{one-to-one} mapping does, the models with and without DQE have very close performance. That is to say, without training disturbance, DQE does not benefit DCA's tolerance to the calibration error. Such fact suggests that when training data fail to cover the gaps between the initial projection and the optimal alignments during validation, the initial image features in DQE provide no valid information for the offset prediction. The experimental results are reported in the supplementary. 2) The boundary of more fusion models' tolerance to the calibration error could be explored in the future.

\section{Conclusion}
In this paper, we propose a Dynamic Cross Attention (DCA) module that adopts a novel \textit{one-to-many} mapping to model the dynamic calibration-based alignment of point clouds and images. DCA learns multiple offsets from the initial projection towards the neighborhood and the adjacent feature levels and thus can tolerate a level of calibration error. The \textit{dynamic query enhancement}(DQE) is proposed to incorporate the initial feature pairs into the offset prediction, which further strengthens DCA's tolerance to the initial misalignment. The whole fusion architecture DCAN adopts multi-level image features and adapts to multiple representations of point clouds, enabling DCA as a plug-in fusion module. Extensive experiments prove DCA's generalized effectiveness and its robustness against the calibration error.



%
%
\bibliographystyle{splncs04}
\bibliography{egbib}
\end{document}